\pdfoutput=1

\documentclass[11pt]{article}

\usepackage[final]{acl}

\usepackage{times}
\usepackage{latexsym}
\usepackage{multirow}
\usepackage{hyperref}

\usepackage[T1]{fontenc}

\usepackage[utf8]{inputenc}

\usepackage{microtype}

\usepackage{inconsolata}

\usepackage{amsmath}
\usepackage{bm}
\usepackage{mathtools}
\usepackage{amsfonts}
\usepackage{booktabs}

\title{Diffusion Guided Language Modeling}

\author{Justin Lovelace\thanks{Correspondence to <\href{mailto:jl3353@cornell.edu}{jl3353@cornell.edu}>.} \quad
  Varsha Kishore \quad
  Yiwei Chen \quad
  Kilian Q. Weinberger \\
  Cornell University\\
}
\newcommand{\method}{DGLM}
\newcommand{\methodfull}{Diffusion Guided Language Modeling}

\begin{document}
\maketitle
\begin{abstract}
Current language models demonstrate remarkable proficiency in text generation. However, for many applications it is desirable to control attributes, such as sentiment, or toxicity, of the generated language---ideally tailored towards each specific use case and target audience.
For auto-regressive language models, existing guidance methods are prone to decoding errors that cascade during generation and degrade performance.
In contrast, text diffusion models can easily be guided with, for example, a simple linear sentiment classifier---however they do suffer from significantly higher perplexity than auto-regressive alternatives. 
In this paper we use a guided diffusion model to produce a latent proposal that steers an auto-regressive language model to generate text with desired properties. Our model inherits the unmatched fluency of the auto-regressive approach and the plug-and-play flexibility of diffusion. 
We show that it outperforms previous plug-and-play guidance methods across a wide range of benchmark data sets. Further, controlling a new attribute in our framework is reduced to training a single logistic regression classifier.  Our code is available at \url{https://github.com/justinlovelace/Diffusion-Guided-LM}.

\end{abstract}

\section{Introduction}


The rapid and ubiquitous adoption of (large) language models (LMs) raises a critical parallel challenge:  how do we effectively guide their generation to be safe and fitting for each application and target audience? For example, one might want an LM to use different language if it interacts with kindergarteners, writes a comedy sketch, provides legal support, or summarizes news articles.

\begin{figure}[h]
\centering
\includegraphics[width=.6\linewidth]{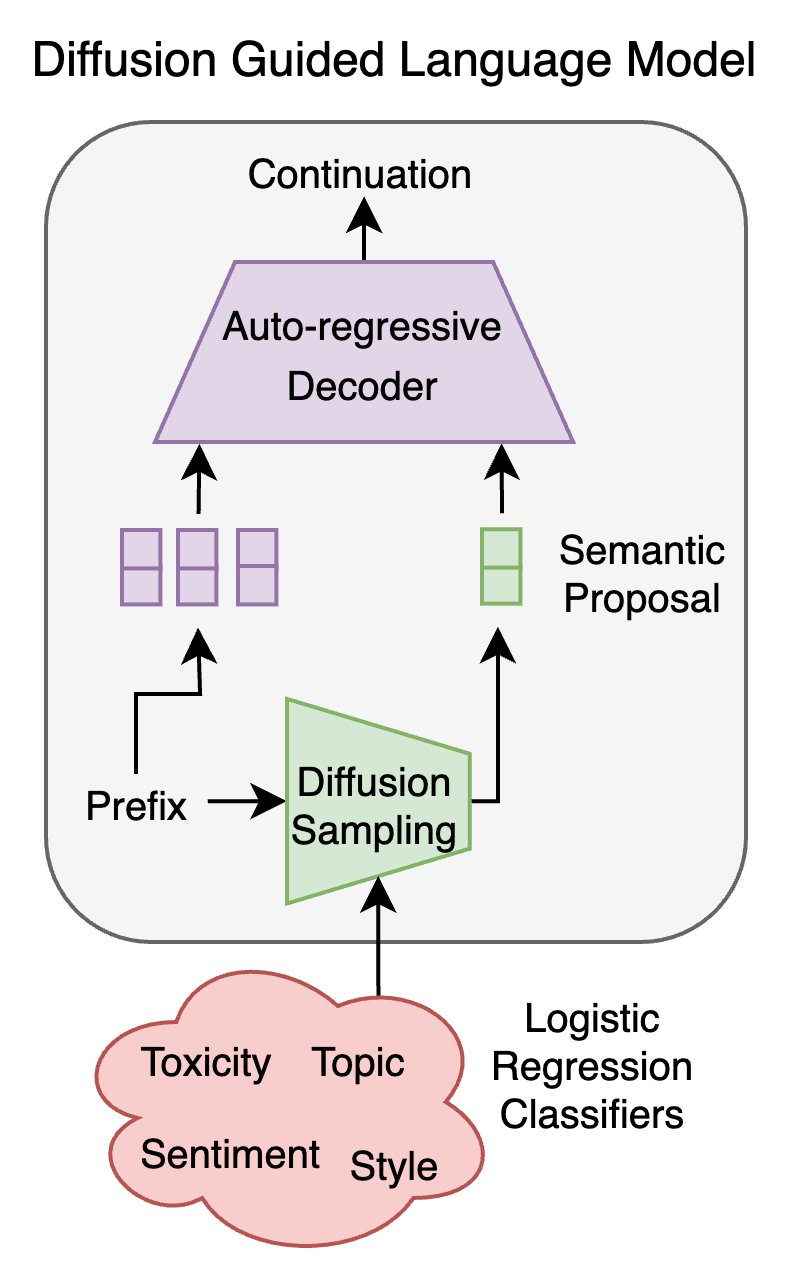}
\caption{Illustration of our Diffusion Guided Languauge Model. We pre-train the autoregressive decoder and the diffusion network used to generate semantic proposals. During generation, we can perform plug-and-play control with simple, linear attribute classifiers.}
\label{fig:intro_overview}
\end{figure}

Currently, the most successful LLM paradigm is to train a single large auto-regressive model that can be used for many tasks~\cite{raffel2020exploring,brown2020language}.
Different approaches to guide generation of such LLMs exist, each with their own strengths and weaknesses. 
A popular way to control the generation is to align the LM through fine-tuning. These approaches are very effective, but as they change the actual model weights, they can deteriorate the LM's performance \citep{lazaridou2020multi, ouyang2022training, bubeck2023sparks, noukhovitch_language_2023}. Further, if new applications require a unique combination of attribute preferences (e.g. \textit{humorous} but  \textit{not toxic}), new dedicated models must be fine-tuned and hosted. 
In contrast, plug and play approaches do not change the model weights and instead utilize additional light-weight classifiers or heuristics to influence the generation process~\cite{dathathri2019plug,yang2021fudge,krause2021gedi,liu2021dexperts,deng2023reward}. Such approaches are highly flexible and do not require fine-tuning or the hosting of dedicated models. However, as they typically alter the logits in the final layer, they are prone to creating decoding errors that cascade through the auto-regressive generation process and deteriorate the output quality. 

One alternative to auto-regressive generation is provided by 
diffusion models~\cite{diffusion_orig, song2020score,ho2020denoising}. Originally gaining prominence in image generation, diffusion models learn to iteratively ``denoise'' samples of Gaussian noise into samples from a target data distribution (e.g. natural images, or text completions). 

Crucially, this iterative generation  naturally allows for plug-and-play control through a simple likelihood function \citep{dhariwal2021diffusion}. Minor errors introduced by the  guidance mechanism can be corrected by the diffusion model later in the generative process. Pre-trained image diffusion models, for instance, can incorporate plug-and-play guidance at inference-time to perform tasks such as super-resolution and in-painting, without any task-specific training. 

Recent work has begun to explore the application of diffusion to the discrete problem of language generation~\cite{li2022diffusion,gong2022diffuseq,lovelace2023latent,gulrajani2023likelihood,zhang2023planner}. Diffusion language models have demonstrated positive results in controllable generation, but still exhibit poor perplexity and generation quality compared to auto-regressive models.

In this paper we propose a novel framework, \emph{\methodfull{} (\method{})}, (see \autoref{fig:intro_overview}) that integrates the fluency of auto-regressive generation with the flexibility of continuous diffusion. 
We develop a diffusion network that, given some text prefix, generates continuous semantic proposals of language continuations. 
These semantic proposals act as soft prompts and guide a fluent auto-regressive model to generate language aligned with the proposal. During pre-training, we condition the language decoder on embedded representations of the ground truth continuation, teaching the decoder that the semantic proposals contain valuable information. During inference time, we let the diffusion model generate its own proposal continuation from the prefix, guided by a simple linear classifier to ensure the desired attributes. The proposal vectors function as additional prompts for the decoder and steer it towards a fluent continuation that inherits the attributes of the proposal.

\method{} has several compelling properties: 1. It decouples model training from attribute control.  2. Controlling a new attribute only requires the training of a simple logistic regression classifier. 3. Empirically, \method{} is extremely effective and outperforms the current state-of-the-art in plug-and-play control across diverse benchmark data sets.

\section{Background: Diffusion Models}

We introduce diffusion models \citep{diffusion_orig, ho2020denoising, song2020score}, following the presentation of \citet{kingma2023understanding} most closely. Given some dataset drawn from an unknown distribution $q(\mathbf{x})$, our goal is to learn a generative model $p_\theta(\mathbf{x})$, shorthanded as $p(\mathbf{x})$, that approximates the unknown data distribution $q(\mathbf{x})$. The observed data $\mathbf{x}$ could be an image, text, or some latent feature vector \citep{rombach2021highresolution}. 

\paragraph{Forward process.}
Diffusion models consist of a forward process and a generative process. The forward process defines a gradual transition from the data distribution to a Guassian distribution. This introduces a series of  increasingly noisy latent variables $\mathbf{z}_t$ for timesteps $t\in[0,1]$ \citep{kingma2021variational}. This Gaussian diffusion process defines the conditional distribution $q(\mathbf{z}_{0,\ldots,1}|\mathbf{x})$. For every $\mathbf{t}\in[0,1]$, the marginal $q(\mathbf{z}_{t}|\mathbf{x})$ is given by:
\[
\mathbf{z}_t = \alpha_{t} \mathbf{x} + \sigma_{t} \bm{\epsilon}, \quad \text{where} \quad \bm{\epsilon} \sim \mathcal{N}(\mathbf{0}, \mathbf{I})
\]
We utilize the common variance-preserving formulation, where $\sigma_{t}^2=1-\alpha_{t}^2$. The noise level is also commonly written in terms of the log Signal-to-Noise Ratio (SNR), ${\lambda_t = \log \alpha_{t}^2/\sigma_{t}^2}$. The noise schedule, specified by $\alpha_t\in [0,1]$, is a strictly monotonically decreasing function defined so that the process starts with the original input, $\mathbf{z}_0 \approx \mathbf{x}$, and the final latent becomes approximately Gaussian, $q(\mathbf{z}_1) \approx \mathcal{N}(\mathbf{z}_1; \mathbf{0}, \mathbf{I})$.

\paragraph{Generative model.}
The generative process reverses the forward process, defining a gradual transition from Gaussian noise to the data distribution. The generative model defines a probability distribution over the latent variables, $p(\mathbf{z}_0,\ldots,\mathbf{z}_1)$.

Given access to the score function $\nabla_{\mathbf{z}
}\log q_t(\mathbf{z}
)$,
 the gradient of the log probability density function, the forward process can be reversed exactly. 
 Diffusion models learn to approximate the score function with a neural network, $\mathbf{s}_\theta(\mathbf{z};\lambda)\approx\nabla_{\mathbf{z}}\log q_t(\mathbf{z})$, and use the estimated score function to approximately reverse the forward process. If $\mathbf{s}_\theta(\mathbf{z};\lambda)\approx\nabla_{\mathbf{z}}\log q_t(\mathbf{z})$, then our generative distribution is close to the true distribution. 
 
 This enables us to draw samples from a Guassian distribution $\mathbf{z}_1 \sim p(\mathbf{z}_1)$, and approximately solve the reverse diffusion process using the estimated score $\mathbf{s}_\theta(\mathbf{z};\lambda)$. In this work, we use the standard DDPM sampler from \citet{ho2020denoising}.

 \paragraph{Training objective.} \citet{song2019generative} showed that score networks, $\mathbf{s}_\theta(\mathbf{z};\lambda)$, can be learned with a denoising score matching (DSM) loss over all data points $\mathbf{x}\sim\mathcal{D}$ and noise levels:
\begin{align*}
    &\mathcal{L}_{\text{DSM}}(\mathbf{x}) = \\
    &\mathbb{E}_{t,\mathbf{x}, \bm{\epsilon}}[{w}(\lambda_t)\cdot \lVert\mathbf{s}_\theta(\mathbf{z}_t;\lambda) - \nabla_{\mathbf{z}_t}\log q(\mathbf{z}_t|\mathbf{x})\rVert_2^2],
\end{align*}
where $w(\lambda_t)$ is a SNR-dependent weighting term that is tuned to emphasize noise levels important for downstream sample quality.

The neural network can be parameterized in terms of the noise ($\bm{\epsilon}$), the data ($\mathbf{x}$), or the velocity ($\mathbf{v}\coloneq \alpha_{t} \mathbf{x} + \sigma_{t}\bm{\epsilon}$) \citep{salimans2022progressive} because of the following relationship: 
\begin{align*}
    \nabla_{\mathbf{z}_t}\log q(\mathbf{z}_t|\mathbf{x}) &= -\bm{\epsilon}/\sigma_{t}\\
    &=-\sigma_{t}^{-2}(\mathbf{z}_{t}-\alpha_{t}\mathbf{x})\\
    &=-\mathbf{z}_{t} - (\alpha_{t}/\sigma_{t})\mathbf{v}.
\end{align*} 
In practice, people have found that parameterizing the neural network as an $\bm{\epsilon}$-prediction or a $\mathbf{v}$-prediction model improves training stability and downstream performance \citep{ho2020denoising, salimans2022progressive}. We follow the best practices established in recent work \citep{kingma2023understanding} and adopt the $\mathbf{v}$-parameterization:
\begin{align*}
    &\mathcal{L}_{\mathbf{v}}(\mathbf{x}) = \mathbb{E}_{t,\mathbf{x}, \bm{\epsilon}}[{w}(\lambda_t)\cdot \lVert\hat{\mathbf{v}}_\theta(\mathbf{z}_t;\lambda) - \mathbf{v}_t\rVert_2^2].
\end{align*}
The above relationships also mean that at every timestep, $t$, the diffusion network provides us with the minimum mean-squared error (MMSE) estimate of the clean data:
\[\hat{\mathbf{x}}_\theta(\mathbf{z}_t, \lambda_t)= \alpha_t\mathbf{z}_t - \sigma_t \hat{\mathbf{v}}_\theta(\mathbf{z}_t;\lambda).\]
\begin{figure}[h]
\centering
\includegraphics[width=1\linewidth]{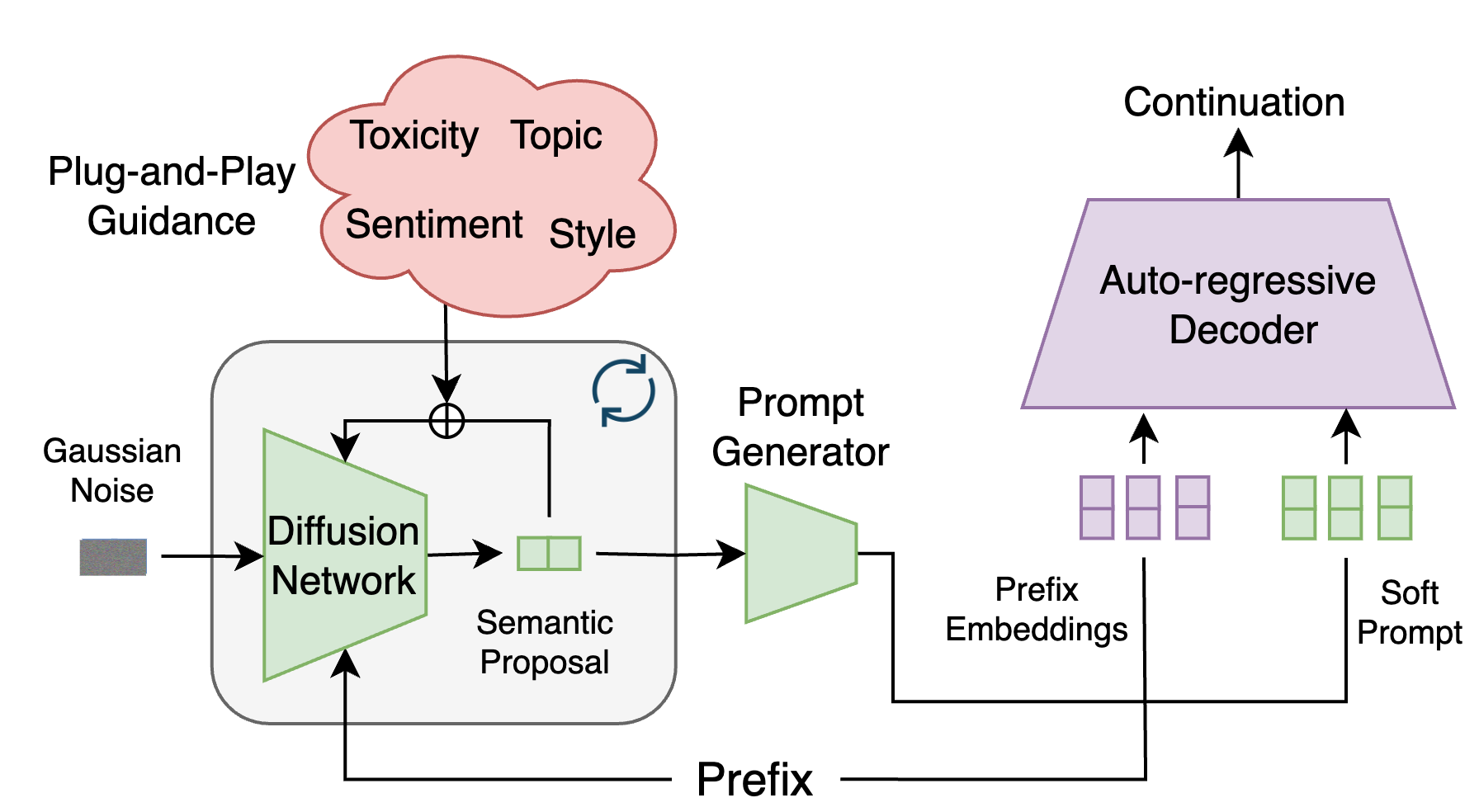}
\caption{Overview of our full generation pipeline. Given some prefix, we first generate an embedded representation of the language continuation with a diffusion model. During this stage, we can optionally intervene with a lightweight classifier for plug-and-play guidance. We map the continuation embedding to a soft prompt to guide an auto-regressive decoder to generate language aligned with the semantics of the generated embedding.}
\label{fig:overview}
\end{figure}

\paragraph{Plug-and-play control.}
When drawing samples $\mathbf{x}$, we want them to meet certain conditional criteria $\mathbf{y}$ such as a class label. One can learn the conditional score function $\nabla_{\mathbf{z}_t} \log p_t(\mathbf{z}_t|\mathbf{y})$ directly with a conditional diffusion model. However, like learning a conditional auto-regressive model, this would require a large corpus of annotated data and the conditional model could not be easily adapted to other conditions. We can instead use Bayes' rule to decompose the conditional score at time $t$ into the unconditional score and a likelihood term: 
\begin{align*}
    &\nabla_{\mathbf{z}_t} \log p_t(\mathbf{z}_t|\mathbf{y}) \\
    &=\nabla_{\mathbf{z}_t} \log p_t(\mathbf{z}_t) + \nabla_{\mathbf{z}_t} \log p_t(\mathbf{y}|\mathbf{z}_t).
\end{align*}

This decomposition shows that we can perform conditional generation with an unconditional diffusion model if we can estimate $\nabla_{\mathbf{z}_t} \log p_t(\mathbf{y}|\mathbf{z}_t)$, the gradient of the log-likelihood of the condition given the latent \citep{dhariwal2021diffusion}. 
 
Diffusion Posterior Sampling (DPS) \citep{chung2023diffusion} utilizes a conditional distribution over noiseless data $p(\mathbf{y}|\mathbf{x})$ and the MMSE estimator $\hat{\mathbf{x}}_\theta(\mathbf{z}_t, \lambda_t)$ to approximate the conditional: \[\nabla_{\mathbf{z}_t} \log p_t(\mathbf{y}|\mathbf{z}_t) \approx \nabla_{\mathbf{z}_t} \log p(\mathbf{y}|\hat{\mathbf{x}}_\theta(\mathbf{z}_t, \lambda_t)).\] 
If the distribution of noiseless data $p(\mathbf{y}|\mathbf{x})$ is differentiable with respect to $\mathbf{x}$, the DPS approximation is differentiable with respect to $\mathbf{z}_t$. We can therefore utilize a lightweight classifier over clean data to guide an unconditional diffusion model to sample data $\mathbf{x}$ consistent with some criteria $\mathbf{y}$ in a plug-and-play manner.

In practice, people often introduce some guidance weight term $s$ as a hyperparameter
\begin{align*}
    &\nabla_{\mathbf{z}_t} \log p_t(\mathbf{z}_t|\mathbf{y}) \\
    &=\nabla_{\mathbf{z}_t} \log p_t(\mathbf{z}_t) + s\cdot \nabla_{\mathbf{z}_t} \log p_t(\mathbf{y}|\mathbf{z}_t),
\end{align*}
where setting $s>1.0$ increases the influence of the conditional information. This can be viewed as sampling from a modified distribution ${\tilde{p}_t(\mathbf{z}_t|\mathbf{y}) 
\propto p_t(\mathbf{z}_t)p_t(\mathbf{y}|\mathbf{z}_t)^s}$.

\section{Diffusion Guided Language Modeling}
We present an overview of our framework in \autoref{fig:overview}. Our method has three main components---a diffusion network, a lightweight prompt generator, and a pre-trained auto-regressive decoder. Given some textual prefix, we use the diffusion model to sample an embedded, semantic proposal of a possible continuation. During sampling, we can optionally perform plug-and-play control to enforce some condition (e.g. low toxicity). After sampling the semantic embedding, the prompt generator is used to process the embedding into a soft prompt, which then guides the auto-regressive decoder to generate text aligned with the proposal.

\subsection{Semantic Proposal Conditioning} 
Sentence-T5 \citep{ni2022sentence} is a sentence encoder that is trained contrastively, producing embeddings that capture high-level semantics while being robust to shallow surface-form variations. Because of these properties, we learn our diffusion model in its latent space to generate semantic proposals\footnote{We use Sentence-T5-XL in this work.}.

\begin{figure}[h]
\centering
\includegraphics[width=0.6\linewidth]{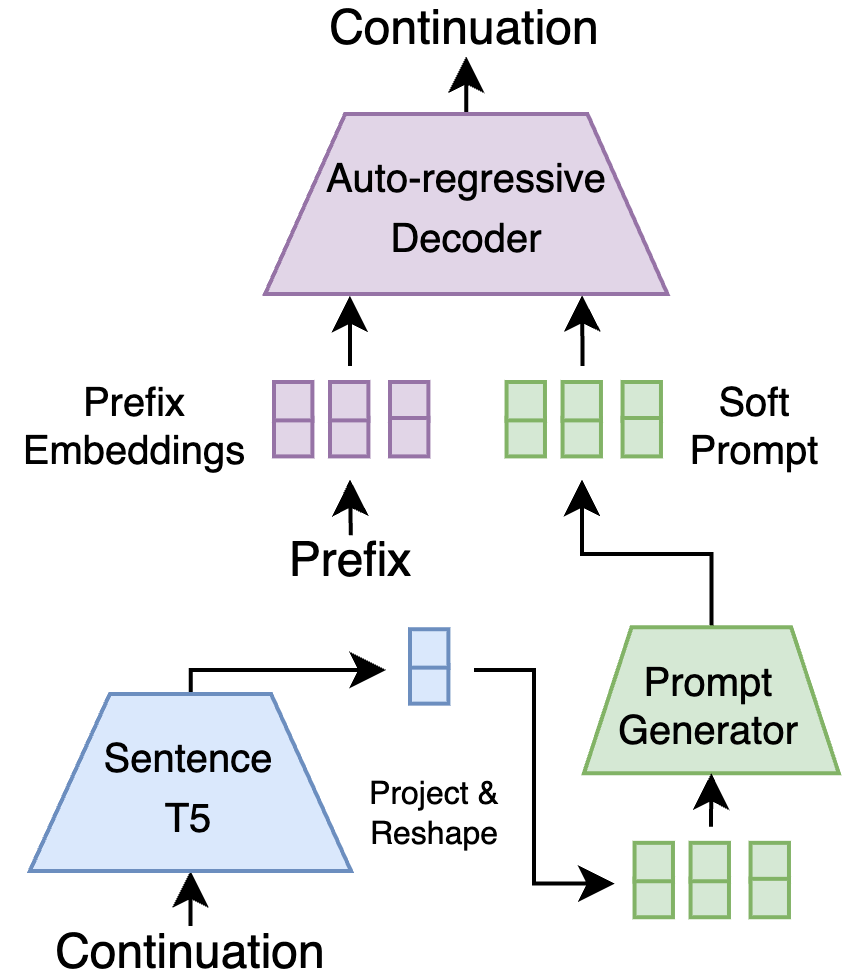}
\caption{Overview of our semantic conditioning stage.}
\label{fig:prompt_generator}
\end{figure}

In order to condition the auto-regressive decoder on embeddings from Sentence-T5, we introduce a lightweight prompt generator that maps the embedding to a soft prompt for the decoder (see \autoref{fig:prompt_generator}). We fine-tune the prompt generator and decoder to generate continuations that correspond to the embeddings from the frozen Sentence-T5 encoder.

Given some text sequence, we split it to obtain a prefix and continuation. We use Sentence-T5 to embed the continuation into a 768-dimensional vector, denoted $\mathbf{x}_{\text{cont}}$. The prompt generator linearly projects the embedding to dimension $4d$, splits it into $k=8$ feature vectors, and then further refines them using a small transformer~\cite{morris2023text}. This yields a sequence of $k$ soft tokens that guide the auto-regressive model to reconstruct the continuation. The input training sequence therefore consists of the prefix text and the soft prompt, which are used to predict the text continuation with teacher forcing. 

The auto-regressive model is trained with the standard language modeling loss. We mask out the predictions corresponding to the soft tokens from the loss function.
Because the sentence embedding corresponds to the ground-truth continuation, the auto-regressive network will learn to generate text aligned with the Sentence-T5-XL embedding.

\paragraph{Gaussian noise conditioning.} 
During generation, we will be utilizing latent proposals from our diffusion network. While an effective diffusion model produces high-quality proposals, it is difficult to match the quality of the ground-truth embeddings used during pre-training. To improve the robustness of the auto-regressive decoder to minor errors introduced by the diffusion network, we incorporate Gaussian noise augmentation, a technique introduced for cascaded image diffusion models \citep{ho2022cascaded, imagen}.

The prompt generator receives a latent sampled from the forward diffusion process ${\mathbf{z}_t = \alpha_{t} \mathbf{x}_{\text{cont}} + \sigma_{t} \bm{\epsilon}}$, where the noise level is sampled according to some schedule $\alpha_t\in[0,1]$. We also condition the prompt generator on the level of noise. The noise level dynamically adjusts the influence of the proposal embedding on the auto-regressive decoder's output. At low noise levels the decoder relies heavily on the proposal embedding, while at high noise levels, the decoder falls back to standard auto-regressive generation.

During generation, we pass a proposal embedding with some low, but non-negligible, level of noise (we set $\sigma_{t}^2=0.05$ by default) and the auto-regressive decoder will generate text aligned with the proposal while correcting for minor errors introduced by the diffusion network. This also provides us with a knob to tailor the influence of the diffusion network to the application. We report full implementation details in \autoref{tab:autoregressive_hyperparams}.

\subsection{Semantic Diffusion} Our semantic diffusion model operates in the latent space of Sentence-T5, iteratively generating potential text continuations guided by a text prefix. Given a text sequence, we split it into a prefix and a continuation and embed both using Sentence-T5, denoted as $\mathbf{x}_{\text{pref}}$ and $\mathbf{x}_{\text{cont}}$ respectively. 

We train the score network to recover the noisy continuation embedding given the prefix embedding. More formally, the noisy latent is given as $\mathbf{z}_t = \alpha_{t} \mathbf{x}_{\text{cont}} + \sigma_{t} \bm{\epsilon}$ and we parameterize our score network as $\mathbf{s}_\theta(\mathbf{z}_t;\lambda;\mathbf{x}_{\text{pref}})$. We therefore learn to sample from the distribution of possible continuation embeddings for the text prefix.

\begin{figure}[h]
\centering
\includegraphics[width=0.8\linewidth]{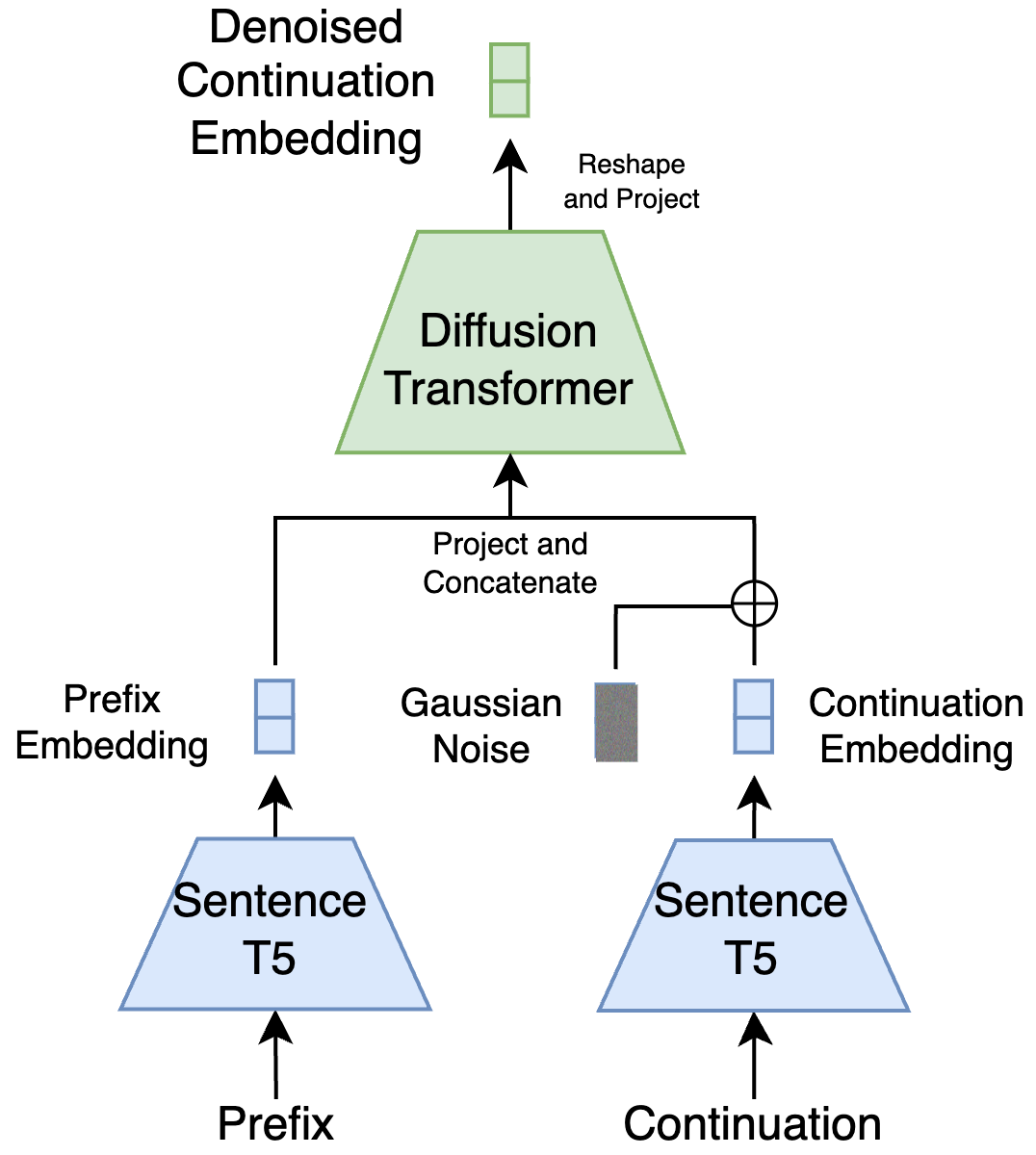}
\caption{Architecture of our diffusion network.}
\label{fig:semantic_diffusion}
\end{figure}

For the diffusion network, we employ a transformer model (see \autoref{fig:semantic_diffusion}).   To prepare the input, we first independently project the noisy latent and prefix embeddings, then split each into 64 feature vectors. We concatenate these element-wise along the feature dimension, giving us 64 representations that we then process with the transformer.

We convert the transformer's output to a single feature vector by inverting the initial projection operation. We down-project and concatenate the 64 feature vectors to create the final vector used for score regression. 
During training, we mask the prompt embedding ($p=0.1$), by replacing it with a learnable null embedding, for classifier-free guidance \citep{ho2022classifier}. This jointly trains unconditional ($\mathbf{s}_\theta(\mathbf{z}_t;\lambda_t)$) and conditional ($\mathbf{s}_\theta(\mathbf{z}_t;\lambda_t;\mathbf{x}_{\text{pref}})$) diffusion networks.
During sampling, we can use guidance weight $w$ to blend predictions as 
\[\tilde{\mathbf{s}}_t = w \hat{\mathbf{s}}_\theta(\mathbf{z}_t;\lambda_t;\mathbf{x}_{\text{pref}}) + (1-w) \hat{\mathbf{s}}_\theta(\mathbf{z}_t;\lambda_t).\]
Setting $w=1.0$ yields the conditional model while setting $w>1.0$ strengthens the influence of the conditioning information and emphasizes prompt-adherent continuations. We report full implementation details of our diffusion model in \autoref{tab:diffusion_hyperparams}.

\begin{table*}[h]
\small
\centering
\resizebox{.8\linewidth}{!}{
\begin{tabular}{lcccccccc}
\toprule
 & & \multicolumn{3}{c}{\textbf{C4}} & \multicolumn{3}{c}{\textbf{OpenWebText}} \\
 \cmidrule(lr){3-5} \cmidrule(lr){6-8}

   & Prefix Guidance ($w$) & $\text{MAUVE}$ $\uparrow$  & OLMo Ppl $\downarrow$  & Div $\uparrow$ &  $\text{MAUVE}$ $\uparrow$  &  OLMo Ppl $\downarrow$  & Div $\uparrow$  \\
  
\midrule
  Reference & - & - & $19.2$ & $58.4$ & - & $17.2$ & $57.6$ \\
  \midrule
  GPT-2$_{\text{Large}}$  & - & $83.9_{.3}$ & $116.3$ & $50.3$ & $88.2_{.3}$ & $17.6$ & $49.2$\\
  \midrule
  \method{}  & 1.0 &  $84.0_{.4}$ & $30.1$ & $50.8$ &  $78.6_{.4}$ & $22.9$ & $50.2$\\
  \method{}  & 1.5 & $85.6_{.4}$ & $23.0$ & $52.5$ &  $82.8_{.3}$ & $17.1$ & $51.4$ \\
  \method{}  & 2.0 & $84.8_{.8}$ & $21.4$ & $53.3$ &  $83.1_{.3}$ & $15.4$ & $52.1$\\
  \method{}  & 2.5 & $84.8_{.1}$ & $20.2$ & $54.0$ &  $83.7_{.4}$ & $15.0$ & $52.4$\\
  \method{}  & 3.0 & $86.6_{.2}$ & $19.8$ & $54.0$ &  $84.5_{.4}$ & $14.7$ & $52.5$\\
  \method{}  & 5.0 & $85.6_{.4}$ & $19.4$ & $53.9$&  $84.0_{.3}$ & $14.2$ & $52.6$\\
\bottomrule
\end{tabular}
}
\caption{Language generation evaluation. For the MAUVE score, we report the standard error of the mean over 5 random seeds.}\label{tab:unconditional}

\end{table*}

\subsection{Plug and Play Control}

To effectively control text generation with desired conditions (denoted as $\mathbf{y}$), we develop a plug-and-play approach leveraging the semantic structure of Sentence-T5's embeddings. We now present the mathematical formulation of our approach.

Our semantic diffusion model estimates the score of possible text continuations within the Sentence-T5 latent space given some prefix: $\nabla_{\mathbf{z}_t} \log p_t(\mathbf{z}_t|\mathbf{x}_{\text{pref}})$. Given some condition $\mathbf{y}$ that we wish to enforce for our sample $\mathbf{x}_{\text{cont}}$ at inference time, we decompose the conditional score using Bayes' rule and the DPS approximation, $\hat{\mathbf{x}}_\theta(\mathbf{z}_t, \lambda_t, \mathbf{x}_{\text{pref}})$, as 
\begin{align*}
    &\nabla_{\mathbf{z}_t} \log p_t(\mathbf{z}_t|\mathbf{x}_{\text{pref}},\mathbf{y}) \\
    &=\nabla_{\mathbf{z}_t} \log p_t(\mathbf{z}_t|\mathbf{x}_{\text{pref}}) + \nabla_{\mathbf{z}_t} \log p_t(\mathbf{y}|\mathbf{z}_t, \mathbf{x}_{\text{pref}})\\
    &\approx \nabla_{\mathbf{z}_t} \log p_t(\mathbf{z}_t|\mathbf{x}_{\text{pref}}) \\
    &+ \nabla_{\mathbf{z}_t} \log p(\mathbf{y}|\hat{\mathbf{x}}_\theta(\mathbf{z}_t, \lambda_t, \mathbf{x}_{\text{pref}}), \mathbf{x}_{\text{pref}}).
\end{align*}

 Since $\mathbf{y}$ depends solely on the continuation and the DPS estimate already incorporates information from the prefix, we assume conditional independence between $\mathbf{y}$ and $\mathbf{x}_{\text{pref}}$ given $\hat{\mathbf{x}}_\theta(\mathbf{z}_t, \lambda_t, \mathbf{x}_{\text{pref}})$. Mathematically, this is expressed as: 
\begin{align*}
\nabla_{\mathbf{z}_t} \log p(\mathbf{y}|\hat{\mathbf{x}}_\theta(\mathbf{z}_t, \lambda_t, \mathbf{x}_{\text{pref}}), \mathbf{x}_{\text{pref}}) \\
\approx \nabla_{\mathbf{z}_t} \log p(\mathbf{y}|\hat{\mathbf{x}}_\theta(\mathbf{z}_t, \lambda_t, \mathbf{x}_{\text{pref}})).
\end{align*}

This simplification allows us to express the conditional score function:
\begin{align*}
    &\nabla_{\mathbf{z}_t} \log p_t(\mathbf{y}|\mathbf{x}_{\text{pref}},\mathbf{z}_t) \\
&\approx \nabla_{\mathbf{z}_t} \log p(\mathbf{y}|\hat{\mathbf{x}}_\theta(\mathbf{z}_t, \lambda_t, \mathbf{x}_{\text{pref}})) \\
&=-\nabla_{\mathbf{z}_t} \ell_\mathbf{y}(\hat{\mathbf{x}}_\theta(\mathbf{z}_t, \lambda_t, \mathbf{x}_{\text{pref}})
\end{align*}
where $\ell_\mathbf{y}(\hat{\mathbf{x}}_\theta(\mathbf{z}_t, \lambda_t, \mathbf{x}_{\text{pref}})$ is the cross-entropy loss. With this, plug-and-play guidance simply requires a classifier within the sentence-T5 latent space.  We employ a linear probe (i.e. logistic regression) in our experiments (see \autoref{app:impl} for additional details). We find that semantic diffusion enables effective control with surprisingly simple classifiers.

\citet{pmlr-v202-song23k} observed that the MMSE estimate, $\hat{\mathbf{x}}_\theta(\mathbf{z}_t, \lambda_t, \mathbf{x}_{\text{pref}})$, introduced approximation errors in the conditional score estimate. They propose sampling around the MMSE estimate \[\hat{\mathbf{x}}^{(i)}\sim\mathcal{N}(\hat{\mathbf{x}}_\theta(\mathbf{z}_t, \lambda_t, \mathbf{x}_{\text{pref}}), \sigma^2_t/\alpha_t^2\mathbf{I}).\] The sampling distribution has large variance early in the sampling process when the DPS estimate is uncertain and converges to the DPS point estimate at the end of the sampling process. They use a Monte-Carlo approach to approximate the guidance with the logmeanexp operation. Adapting this, we compute the guidance term as:  
\[-\nabla_{\mathbf{z}_t} \log(\frac{1}{n}\sum_i^n \exp(\ell_\mathbf{y}(\hat{\mathbf{x}}^{(i)}))).\]

Early in the sampling process, this steers generation towards a region of low loss within the latent space. With $n=32$, using the Monte-Carlo estimate incurs negligible overhead, requiring only 32 logistic regression evaluations.

\section{Datasets and Metrics}

\paragraph{Datasets.} We extract a subset of 10 million instances from C4~\cite{2019t5} to pre-train \method{}. This represents only $2.5\%$ of C4 and scaling the pre-training corpus would likely be fruitful. We follow \citet{geiping2023cramming} and filter out uncompressible text to improve quality. If the number of GPT-2 tokens is more than $t=0.3$ times the raw number of characters, we drop it from the dataset. This removes instances consisting of, for instance, long HTML strings or markdown code.

To evaluate the language generation capabilities of our \method{}, we extract 5000 random validation instances from C4~\cite{2019t5} and OpenWebText~\cite{Gokaslan2019OpenWeb}. We condition the network on the first 32 tokens and generate a 32 token continuation. For our toxicity mitigation experiments, we train our logistic regression model on the Jigsaw Unintended Bias dataset~\cite{jigsaw-unintended-bias-in-toxicity-classification} and evaluate the effectiveness of toxicity mitigation experiments using 5,000 neutral promps from RealToxicityPrompts~\cite{gehmanrealtoxicityprompts}. For our sentiment control experiments, we utilize Amazon Polarity~\footnote{\url{https://huggingface.co/datasets/amazon_polarity}} and SST-2~\cite{socher2013recursive} to train a sentiment classifier, and perform sentiment control using 5,000 neutral prompts from OpenWebText. 

\begin{figure*}[h]
\centering
\begin{minipage}{.4\textwidth}
\centering
\includegraphics[width=\linewidth]{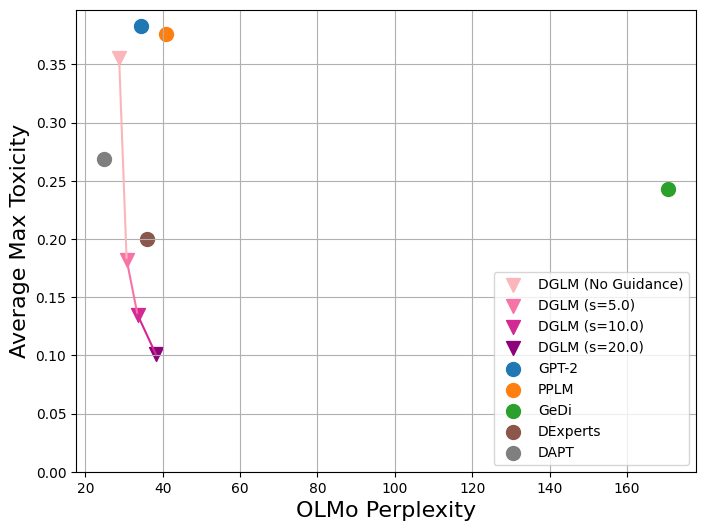}
\end{minipage}%
\begin{minipage}{.4\textwidth}
\centering
\includegraphics[width=\linewidth]{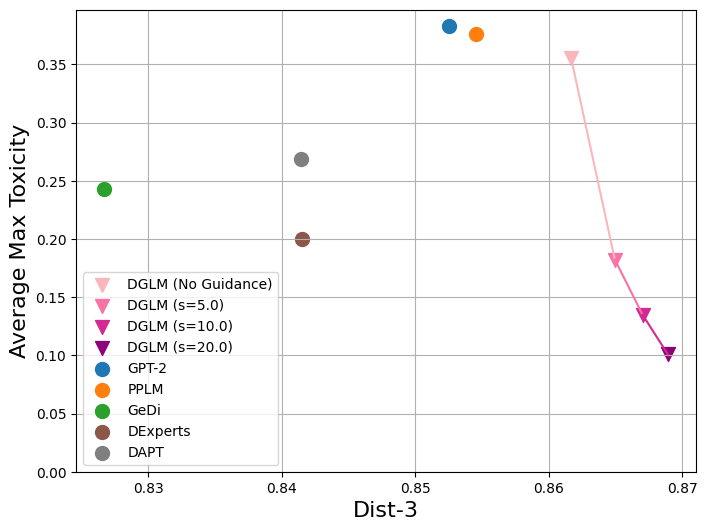}
\end{minipage}
\caption{Effect of mitigating toxicity with increasing guidance weights. Increasing guidance reduces toxicity with minimal loss of fluency. \label{fig:avg_max_toxicity}}
\end{figure*}

\paragraph{Metrics.} We evaluate the fluency of text by measuring its perplexity with the open-source OLMo-1B\footnote{\url{https://huggingface.co/allenai/OLMo-1B}} language model. We also report MAUVE scores~\cite{pillutla2021mauve}, a text generation metric that measures the similarity of generated text with that of reference text using divergence frontiers. To get embeddings for MAUVE, we follow the advice of \citet{he2022blind} and utilize ELECTRA-large \citep{clark2020electra}. To evaluate generation diversity, we use the metric introduced by \citet{su2022contrastive}: ${\text{\textbf{Div}} = \prod_{n=2}^4 \frac{\lvert \text{unique n-grams($\{\mathbf{w}_i\}$)}\rvert}{\lvert \text{total n-grams($\{\mathbf{w}_i\}$)}\rvert}}$ where $\{\mathbf{w}_i\}$ is a set of generated samples.

For the guidance tasks, we generate 25 samples per prompt. We report the OLMo-1B perplexity of the continuations to evaluate the fluency of the generations. We follow prior work and measure the average number of unique 3-grams, denoted Dist-3, in each set of continuations to quantify generation diversity. Along with ensuring that guidance does not degrade language quality or sacrifice diversity, we measure the adherence to the guidance conditions. Following prior work~\cite{deng2023reward,liu2021dexperts}, we use the Perspective API to measure the toxicity of generated text. Because \citet{pozzobon2023on} found that the Perspective API changes significantly over time, we re-score the released generations for all of the baselines with the current version of the API. We measure the \textit{average max toxicity} across 25 generations and the \textit{toxicity rate}, defined as the empirical odds of at least 1 of 25 continuations being classified as toxic.

To evaluate sentiment, we utilize RoBERTa-Large\footnote{\url{https://huggingface.co/siebert/sentiment-roberta-large-english}} \citep{liu2020roberta} fine-tuned on sentiment classification across diverse domains as well as the fine-tuned DistilBERT model \citep{sanh2019distilbert} used by prior work. 

\section{Experimental Results}

\paragraph{Language Generation.}

We validate the effectiveness of our framework on open-ended language generation in \autoref{tab:unconditional} without any plug and play control.
We observe that our method achieves \emph{strong} language generation results, matching or surpassing the reference perplexity with sufficient classifier-free guidance strength. We observe that \method{} leads to consistently \emph{more diverse} generations than the auto-regressive baseline across both datasets. We observed that a handful of very high perplexity samples skews the GPT-2 baseline's perplexity on C4. However, \method{} also achieves stronger MAUVE scores on that dataset.

We examine the impact of Gaussian noise augmentation in \autoref{tab:noise_conditioning}. As an additional metric, we re-embed the generated text with Sentence-T5 and compute the cosine similarity with the proposal embedding\footnote{We follow \citet{bert-score} and rescale the cosine similarity with a baseline computed between random dataset samples.}. We observe that the Gaussian noise augmentation enables the network to smoothly interpolate between auto-regressive generation (low perplexity but poor diversity) and diffusion-guided generation (higher perplexity and diversity). We observe that lower levels of noise montonically improve the decoders adherence to the proposal.

\begin{table}[h]
\small
\centering
\resizebox{\linewidth}{!}{
\begin{tabular}{lccccc}
\toprule
 & & \multicolumn{3}{c}{\textbf{C4}}  \\
 \cmidrule(lr){3-5} 

   & Noise ($\sigma_t^2$) & S-T5 Sim $\uparrow$  & OLMo Ppl $\downarrow$  & Div $\uparrow$ \\
  
\midrule
  Reference & - & $35.7$ & $19.2$ & $58.4$ & \\
  \midrule
  \multirow{8}{*}{\method{}}  & 1.0 &  $36.7$ & $17.3$ & $45.9$  \\
   & 0.8 &  $45.6$ & $21.8$ & $47.1$ \\
   & 0.6 &  $50.9$ & $22.9$ & $48.6$ \\
   & 0.4 &  $54.6$ & $26.1$ & $49.8$ \\
   & 0.2 &  $56.8$ & $28.1$ & $50.3$ \\
   & 0.05 &  $58.5$ & $30.1$ & $50.8$ \\
   & 0.0 &  $59.1$ & $30.7$ & $51.4$ \\
\bottomrule
\end{tabular}
}
\caption{Impact of Gaussian noise augmentation. $\sigma_t^2=1.0$ corresponds to Gaussian noise and $\sigma_t^2=0.0$ corresponds to the clean proposal.}\label{tab:noise_conditioning}

\end{table}

\paragraph{Plug-and-Play Control.}
We utilize \method{} to avoid generating toxic language. We show quantitative results in \autoref{fig:avg_max_toxicity} and \autoref{fig:toxic_rate}. Qualitative examples are presented in \autoref{tab:qualitative}. Plug-and-play guidance with a linear probe effectively mitigates toxicity with negligible trade-offs in fluency. We simultaneously achieve lower perplexity, lower toxicity, and higher diversity than all baselines.

\begin{figure*}
\centering
\begin{minipage}{.4\textwidth}
\centering
\includegraphics[width=\linewidth]{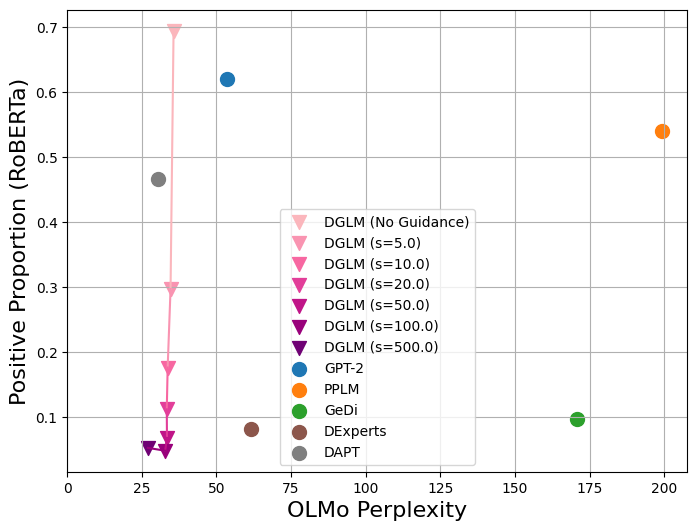}
\end{minipage}%
\begin{minipage}{.4\textwidth}
\centering
\includegraphics[width=\linewidth]{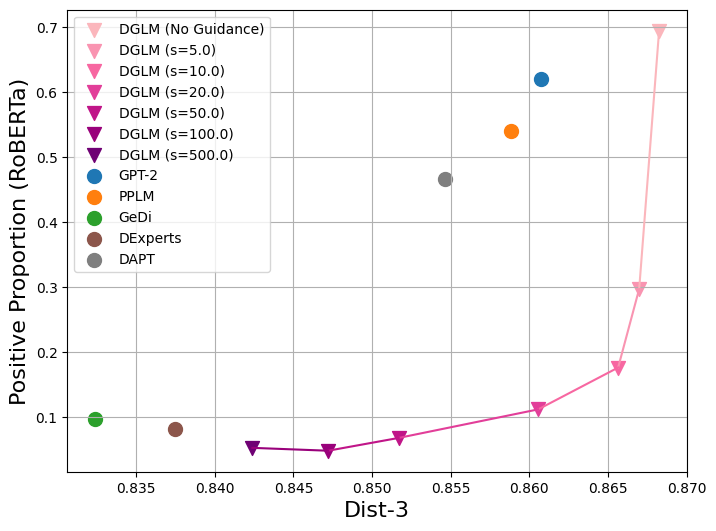}
\end{minipage}
\caption{Effect of guiding generations towards negative sentiment with increasing guidance weights. Increasing guidance improves alignment with the target sentiment while sacrificing diversity.\label{fig:neg_sent}}
\end{figure*}

We also employ \method{} to steer the sentiment of generated text. We present results for negative sentiment in \autoref{fig:neg_sent} and positive sentiment in \autoref{fig:pos_sent}. We observe that our method is similarly effective in this setting, decreasing (or increasing) sentiment with no loss of fluency and minimal loss of diversity for modest guidance values.

\begin{table*}[h]
\small
\centering
\resizebox{\linewidth}{!}{
\begin{tabular}{cccc}
\toprule
 Topic & Sentiment & Prefix & Continuation \\
\midrule
  Sci/ Tech & Negative & Therefore, we will not & \parbox{.52\linewidth}{provide a technical review of the software, including its capabilities, nor will we provide you with any reports or comments regarding the accuracy of information.} \\
  \midrule
  Sports & Positive & Other than that, & \parbox{.52\linewidth}{  I think we really did a great job of letting the fans know how it felt to see them come out in record numbers for an 82
game season.} \\
\bottomrule
\end{tabular}
}
\caption{Language generated by controlling two attributes simultaneously.}\label{tab:compositional_main}

\end{table*}

\paragraph{Compositional Control.}

We present qualitative results from composing multiple attribute classifiers with \method{}. We fine-tune an additional logistic regression model on the AG News topic classification dataset. We then sum the losses for the sentiment and topic classification classifier to guide generation. We find that \method{} successfully enables compositional control and present qualitative examples in \autoref{tab:compositional_main} (additional examples are in \autoref{tab:compositional}). We leave rigorous evaluations of the compositionality of \method{} to future work.

\paragraph{Decoding Overhead.}
Plug-and-play methods for auto-regressive generation often introduce overhead at each decoding step. For example, DExperts employs auxiliary language models that work alongside the primary model. In contrast, \method{} incurs a one-time cost for generating the semantic proposal, which is then amortized across subsequent decoding steps.
We therefore compute runtimes across a range of generation lengths. We report the relative increase in runtime compared to the original GPT-2 model for each method (baseline data from \citet{liu2021dexperts}) in \autoref{tab:time}. As seen in the table, \method{} incurs a large cost for short sequences but has reduced overhead compared to prior methods with modest generation lengths. 

\begin{table}[h]
\small
\centering
\begin{tabular}{lc}
\toprule
  Method & Relative Runtime  \\
  \midrule
   GPT-2 & 1.0x \\
   \midrule
   GeDi & 2.9x \\
   DeXperts (large)&  3.6x \\
   PPLM & 270.1x \\
   \midrule
   \method{} (32 tokens) & 7.4x \\
   \method{} (64 tokens) & 4.4x \\
   \method{} (128 tokens) & 2.6x \\
   \method{} (256 tokens) & 1.7x \\
\bottomrule
\end{tabular}
\caption{Relative runtime compared to GPT-2.}\label{tab:time}
\end{table}

\section{Related Work}
\paragraph{Fine-tuning.} 
Continual pre-training on text from some target domain (domain-adaptive pretraining or DAPT) is an effective technique for controlling attributes in generated text~\cite{gururangan2020don}. \citet{lu2022quark} optimize a reward function by fine-tuning an LM with control tokens for different reward quantiles. Reinforcement Learning with Human Feedback (RLHF) involves training a reward model on human preference data that is then used to fine-tune the LM~\cite{wu2023fine, ouyang2022training}.  \citet{jang2023personalized} train multiple personalized RLFH models and show that these personalized models can be used alone or in conjunction with one another to produce text with desired attributes.

\paragraph{Guided Generation.} Finetuning LMs is expensive and therefore to reduce cost, \citet{dathathri2019plug} proposed Plug and Play Language Model (PPLM), a method that used light-weight classifiers to guide frozen language models during text generation. Similarly, FUDGE~\cite{yang2021fudge} trains classifiers on partial sequences to predict whether a particular attribute is satisfied and updates the output probability distribution accordingly. Instead of using a classifier, GeDi~\cite{krause2021gedi} trains a small class-conditional language model to act as a discriminator and guide the language generation. Similarly, DeXperts~\cite{liu2021dexperts} trains experts and anti-experts by fine-tuning small language models, and using these experts to guide generation. Reward-Augmented Decoding (RAD)~\cite{deng2023reward} trains a reward model to score generations and adjust logit probabilities to promote high-reward tokens.

\section{Conclusion}
We present \methodfull{} (\method{}), a powerful integration of auto-regression and diffusion that enables versatile attribute-guided text generation with lightweight classifiers. The diffusion model generates controllable semantic proposals that guide the language decoder. Extending \method{} to control an unseen attribute only requires learning a single logistic regression model. Experimental results show that \method{} significantly outperforms prior plug-and-play methods, opening avenues for building highly adaptable LMs with user-controllable behavior.

\section{Limitations}
While \method{} demonstrates strong capabilities for guided text generation, we acknowledge important limitations. First, like any system that controls text attributes, it risks potential misuse to steer language in harmful directions. Researchers and practitioners should carefully evaluate generation systems to mitigate these risks.

In addition, \method{} currently has slower inference speed than some plug-and-play baselines when generating short texts (<32 tokens). We expect advances in accelerating diffusion models and distilling diffusion steps will help address this limitation in future work. 

More broadly, while \method{} outperforms recent methods, there is still substantial room for improvement in controllable text generation. The framework currently utilizes simple linear classifiers that may not robustly capture complex attributes. Extending \method{} to complex attributes may require more complex classifiers. We hope our work sparks further research towards reliable and beneficial guided language models.


\section*{Acknowledgements}
This research is supported by grants from the National Science Foundation NSF (IIS-2107161, and IIS-1724282, HDR-2118310), the Cornell Center for Materials Research with funding from the NSF MRSEC program (DMR-1719875), DARPA, arXiv, LinkedIn, and the New York Presbyterian Hospital.

\bibliography{custom, anthology}

\appendix

\section{Additional Figures}
\label{sec:appendix}
\begin{figure*}[h]
\centering
\begin{minipage}{.5\textwidth}
\centering
\includegraphics[width=\linewidth]{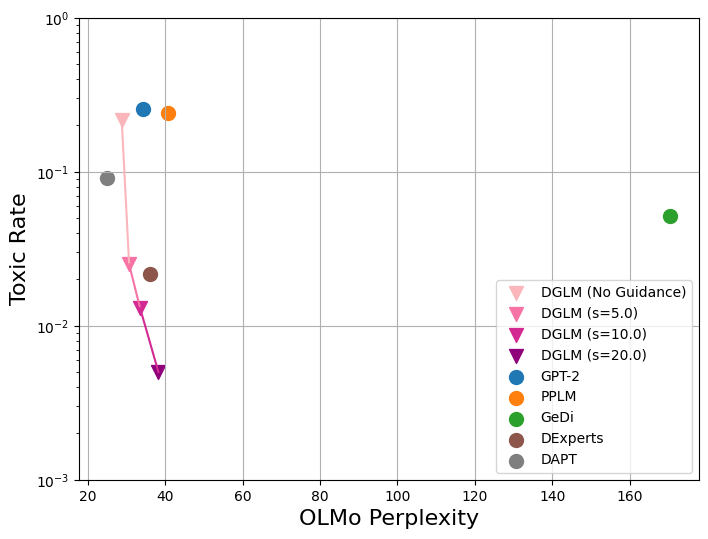}
\end{minipage}%
\begin{minipage}{.5\textwidth}
\centering
\includegraphics[width=\linewidth]{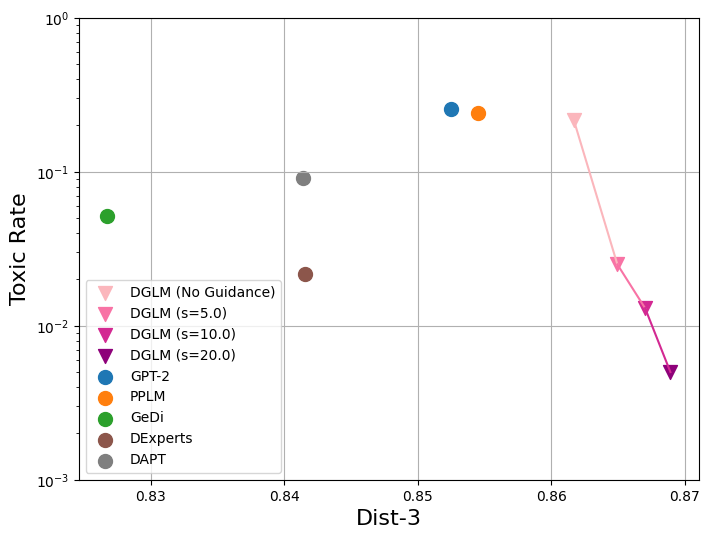}
\end{minipage}
\caption{Effect of plug-and-play toxicity mitigation with increasing guidance weights. We observe that increasing guidance reduces toxicity at the cost of language fluency.\label{fig:toxic_rate}}
\end{figure*}

\begin{figure*}[h]
\centering
\begin{minipage}{.5\textwidth}
\centering
\includegraphics[width=\linewidth]{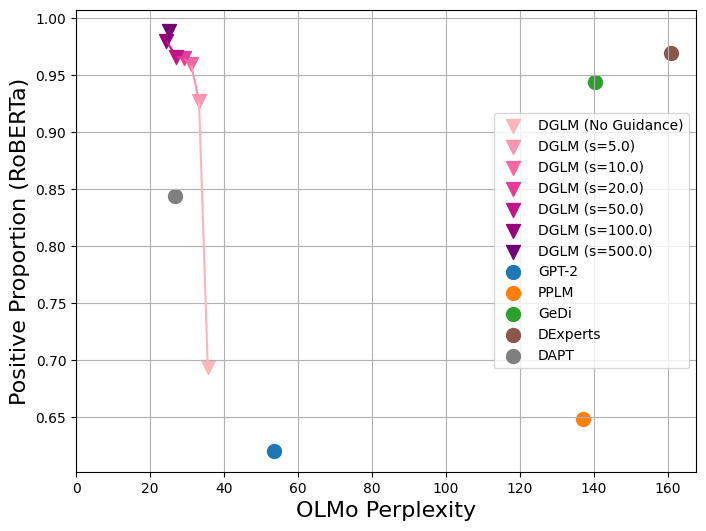}
\end{minipage}%
\begin{minipage}{.5\textwidth}
\centering
\includegraphics[width=\linewidth]{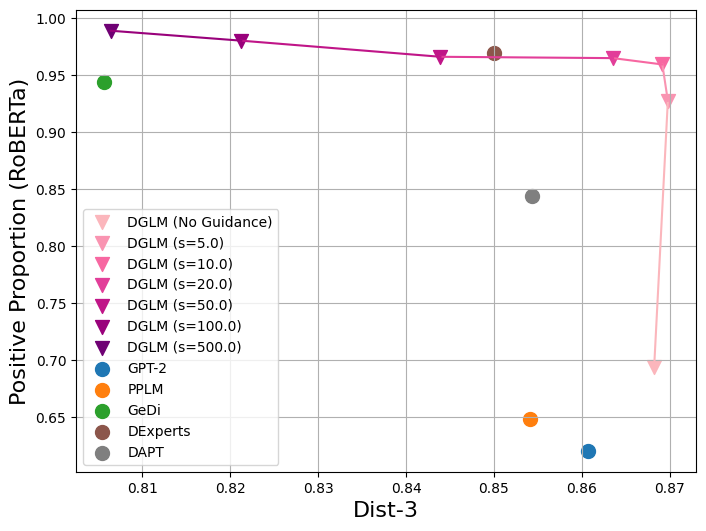}
\end{minipage}
\caption{Effect of guiding generations towards positive sentiment with increasing guidance weights. The left plot shows the impact on language perplexity and the right plot shows the impact on language diversity.\label{fig:pos_sent}}
\end{figure*}

\begin{figure*}[h]
\centering
\begin{minipage}{.5\textwidth}
\centering
\includegraphics[width=\linewidth]{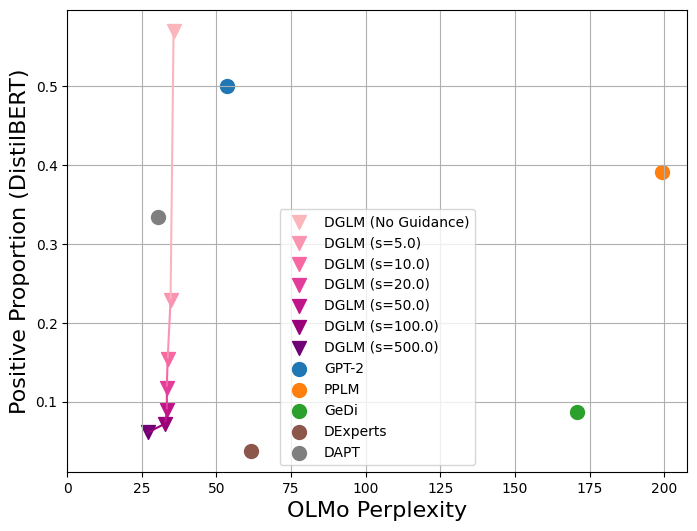}
\end{minipage}%
\begin{minipage}{.5\textwidth}
\centering
\includegraphics[width=\linewidth]{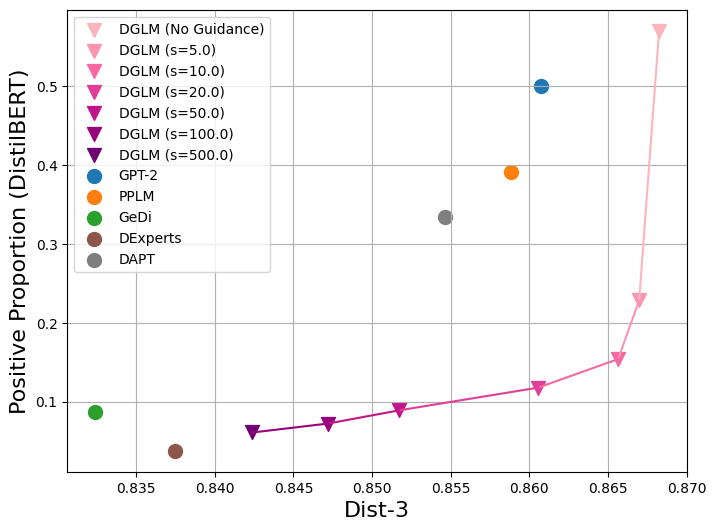}
\end{minipage}
\caption{Effect of guiding generations towards negative sentiment with increasing guidance weights. The left plot shows the impact on language perplexity and the right plot shows the impact on language diversity.\label{fig:distil_neg}}
\end{figure*}

\begin{figure*}[h]
\centering
\begin{minipage}{.5\textwidth}
\centering
\includegraphics[width=\linewidth]{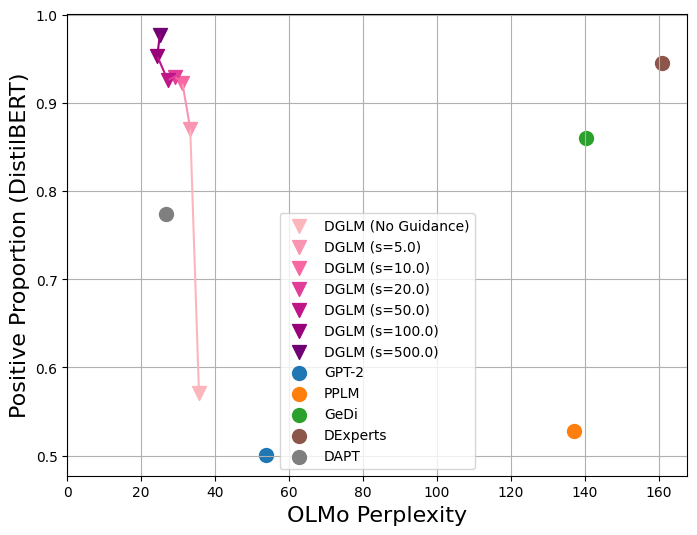}
\end{minipage}%
\begin{minipage}{.5\textwidth}
\centering
\includegraphics[width=\linewidth]{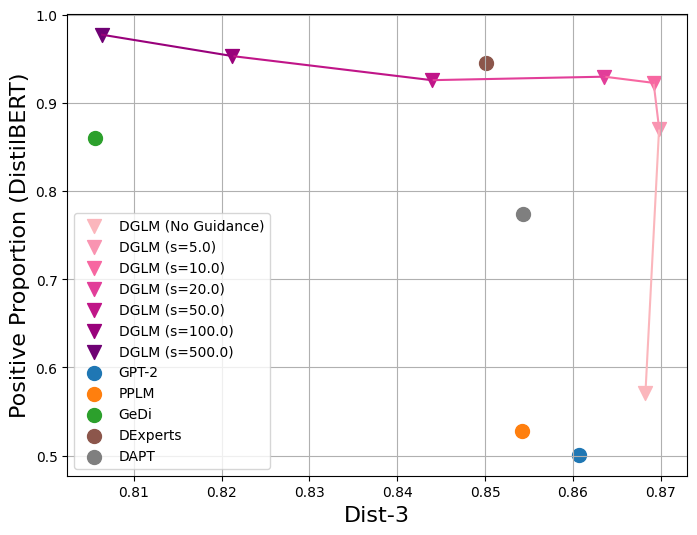}
\end{minipage}
\caption{Effect of guiding generations towards positive sentiment with increasing guidance weights. The left plot shows the impact on language perplexity and the right plot shows the impact on language diversity.\label{fig:distil_pos}}
\end{figure*}

We present toxicity mitigation results with the \textit{Toxic Rate} metric in \autoref{fig:toxic_rate}. We present plug-and-play results with positive sentiment guidance in \autoref{fig:pos_sent}. We present the sentiment guidance results with the DistilBERT classifier used in prior work in \autoref{fig:distil_neg} and \autoref{fig:distil_pos}.

\section{Numerical Results}
We provide the numerical results for our toxicity mitigation and sentiment control experiments in \autoref{tab:toxicity} and \autoref{tab:sentiment}.

For the baseline methods, we observed a handful of extremely high perplexity generations (e.g. >1e4) that significantly increase the average perplexity. Prior work typically filters out these instances when computing the average perplexity\footnote{See \href{https://github.com/alisawuffles/DExperts/blob/4ef198fe4cad76f87f7ceac362171a3bda906303/scripts/evaluation/evaluate\_generations.py\#L31}{here} for an example.}. We did not observe any such high perplexity continuations for our method. We therefore do \emph{not} perform this filtering for our method. 

\begin{table*}[h]
\small
\centering
\resizebox{.8\linewidth}{!}{
\begin{tabular}{lcccc}
\toprule
 Method & Avg. Max Toxicity $\downarrow$  & Toxic Rate $\downarrow$& OLMO Ppl $\downarrow$& Dist-3 $\uparrow$\\  
\midrule
GPT-2 & 0.383 & 0.254 & 34.4 & 0.853 \\
\midrule
DAPT & 0.269 & 0.091 & 24.9 & 0.841 \\

\midrule
PPLM & 0.376 & 0.240 & 40.8 & 0.855 \\
GeDi & 0.243 & 0.051 & 170.5 & 0.827 \\
DExperts & 0.200 & 0.022 & 36.0 & 0.842 \\
\midrule
DGLM (No Guidance) & 0.355 & 0.218 & 28.8 & 0.862 \\
DGLM (s=5.0) & 0.182 & 0.025 & 30.7 & 0.865 \\
DGLM (s=10.0) & 0.135 & 0.013 & 33.5 & 0.867 \\
DGLM (s=20.0) & 0.101 & 0.005 & 38.2 & 0.869 \\

\bottomrule
\end{tabular}
}
\caption{Toxicity Mitigation Results.}\label{tab:toxicity}

\end{table*}

\begin{table*}[h]
\small
\centering
\resizebox{.8\linewidth}{!}{
\begin{tabular}{lcccc}
\toprule
\multicolumn{4}{c}{\textbf{Target Sentiment: Positive}} \\
\midrule
 Method & Positive Prop. (RoBERTa) $\uparrow$ & OLMO Ppl $\downarrow$& Dist-3 $\uparrow$\\  
\midrule
GPT-2 & 0.621 & 53.6 & 0.861 \\
\midrule
DAPT & 0.844 & 26.8 & 0.854 \\
\midrule
PPLM & 0.649 & 137.1 & 0.854 \\
GeDi & 0.944 & 140.4 & 0.806 \\
DExperts & 0.969 & 160.8 & 0.850 \\
\midrule
DGLM (No Guidance) & 0.694 & 35.6 & 0.868 \\
DGLM (Guidance 5.0) & 0.927 & 33.2 & 0.870 \\
DGLM (Guidance 10.0) & 0.959 & 31.1 & 0.869 \\
DGLM (Guidance 20.0) & 0.965 & 29.1 & 0.864 \\
DGLM (Guidance 50.0) & 0.966 & 27.2 & 0.844 \\
DGLM (Guidance 100.0) & 0.980 & 24.3 & 0.821 \\
DGLM (Guidance 500.0) & 0.989 & 25.0 & 0.806 \\
\midrule[\heavyrulewidth]
\multicolumn{4}{c}{\textbf{Target Sentiment: Negative}} \\
\midrule
Method & Positive Prop. (RoBERTa) $\downarrow$ & OLMO Ppl $\downarrow$& Dist-3 $\uparrow$\\  
\midrule
GPT-2 & 0.621 & 53.6 & 0.861 \\
\midrule
DAPT & 0.466 & 30.3 & 0.855 \\
\midrule
PPLM & 0.540 & 199.1 & 0.859 \\
GeDi & 0.097 & 170.7 & 0.832 \\
DExperts & 0.082 & 61.7 & 0.837 \\
\midrule
DGLM (No Guidance) & 0.694 & 35.6 & 0.868 \\
DGLM (Guidance 5.0) & 0.297 & 34.6 & 0.867 \\
DGLM (Guidance 10.0) & 0.176 & 33.6 & 0.866 \\
DGLM (Guidance 20.0) & 0.112 & 33.3 & 0.861 \\
DGLM (Guidance 50.0) & 0.068 & 33.4 & 0.852 \\
DGLM (Guidance 100.0) & 0.048 & 32.8 & 0.847 \\
DGLM (Guidance 500.0) & 0.053 & 27.0 & 0.842 \\
\bottomrule
\end{tabular}
}
\caption{Sentiment Control Results.}\label{tab:sentiment}

\end{table*}

\section{Implementation Details}
\label{app:impl}

We train all of the models in this work on a single NVidia A6000 GPU.

\paragraph{Transformer Implementation.}  We use different configurations of the same transformer architecture for the prompt generator and the diffusion network. We utilize a pre-normalization transformer \citep{vaswani2017transformer, xiong2020layer} with RMSNorm \citep{zhang2019root} and SwiGLU activations \citep{shazeer2020glu}. We condition the transformer on the level of noise by mapping $\alpha_t$ to a sinusoidal positional embedding \citep{vaswani2017transformer} and pass it through an MLP with a single hidden layer to obtain a time embedding. We apply adaptive RMSNorm conditioned on this time embedding before the feedforward layers and attention layer \citep{peebles2022scalable}. We utilize query-key RMSNorm \citep{dehghani2023scaling} for the self-attention mechanisms because it has been shown to improve stability. 

\paragraph{Diffusion Network.}
We employ the $\mathbf{v}$-parameterization and minimize:
\begin{align*}
    &\mathcal{L}_{\mathbf{v}}(\mathbf{x}) = \mathbb{E}_{t,\mathbf{x}, \bm{\epsilon}}[{w}(\lambda_t)\cdot \lVert\hat{\mathbf{v}}_\theta(\mathbf{z}_t;\lambda) - \mathbf{v}_t\rVert_2^2].
\end{align*}
To set the weighting function, we followed the advice of \cite{karras2022elucidating} and parameterized it with a log-normal distribution based on the noise levels where the model was best able to minimize the loss. This led us to set $w(\lambda_t)= \mathcal{N}(\lambda_t;0,2.4)$. Consistent with past work \citep{balaji2022ediffi}, we observed that increasing weights at high noise levels improved the alignment of generations with the conditioning information. For our final weighting function, we therefore used a fat-tailed Cauchy distribution for the left half of the distribution and a normal distribution for the right half. This gives us 
\begin{align*}
w(\lambda_t)= \begin{cases}
\frac{1}{Z_\text{c}} \text{Cauchy}(\lambda_t;0, 2.4) & \text{if } \lambda_t < 0 \\
\frac{1}{Z_\text{n}} \mathcal{N}(\lambda_t;0,2.4) & \text{if } \lambda_t \geq 0
\end{cases}
\end{align*}
where $Z_\text{c}$ and $Z_\text{n}$ are normalization constants such that the density of each distribution at 0 is re-scaled to 1. For training, we utilize the adaptive noise scheduler introduced by \citet{kingma2023understanding} to reduce the variance of the loss estimate.

\paragraph{Sampling Configuration.} We use the stochastic DDPM sampler with 50 sampling steps with the cosine noise schedule \citep{improved-ddpm-nichol21a}. We follow \citet{hoogeboom2023simple} and set the variance for the DDPM sampler to a log-scale interpolation between the upper and lower bounds of the variance from \citet{ho2020denoising}: ${\sigma^2 = \exp( v \log(\sigma_{\text{max}}^2)  + (1-v)\log(\sigma_{\text{min}}^2))}$ with $v=0.2$ . We did not explore this choice in detail and further exploration of sampling configurations would likely improve performance.

\paragraph{Logistic Regression Classifiers.}
We train logistic regression models with sci-kit learn. We utilize the default L-BFGS solver with L2 regularization of 1e-3. We use balanced class weights for the toxicity classifier due to the class imbalance in the toxicity dataset. Our toxicity and sentiment classifiers achieve an Area Under the Receiver Operating Curve of 83.7 and 95.7, respectively.

\begin{table*} 
\small
\centering
\caption{Implementation details for auto-regressive pre-training stage.}\label{tab:autoregressive_hyperparams}
\begin{tabular}{lc}
\toprule
  \midrule
  Prompt Generator Architecture & Pre-Activation Transformer \cite{vaswani2017transformer, xiong2020layer} \\
  Soft Prompt Tokens & 8 \\
  Transformer Layers &  6 \\
  Transformer Dimension & 768 \\
  Self-Attention Heads & 12  \\
  Activation Function & SwiGLU \citep{shazeer2020glu} \\
  Normalization Layer & Adaptive RMSNorm \citep{zhang2019root, peebles2022scalable}  \\
  Max Seq Length & 96  \\
  Optimizer & AdamW \citep{loshchilov2018decoupled}\\
  Learning Rate & {5e-6} \\
  ($\beta_1$, $\beta_2$) & {(0.9, 0.99)} \\
  Batch Size &  64 \\
  Warmup Steps & 5000 \\
  Learning Rate Schedule & Cosine Decay \\
  Weight Decay & {.02} \\
  Gradient Clipping & {1.0}\\
  Batch Size & 64 \\
Augmentation Noise Schedule & Scaled Cosine (s=3.0) \cite{hoogeboom2023simple} \\
  Training Steps & {250k} \\
\bottomrule
\end{tabular}
\end{table*}

\begin{table*} 
\small
\centering
\caption{Implementation details for diffusion training.}\label{tab:diffusion_hyperparams}
\begin{tabular}{lc}
\toprule
  \midrule
  Diffusion Architecture & Pre-Activation Transformer \cite{vaswani2017transformer, xiong2020layer} \\
  Input Tokens & 64 \\
  Transformer Layers &  12 \\
  Transformer Dimension & 768 \\
  Transformer Output Projection Dimension & 96 \\
  Final Feature Vector Dimension & 96*64 \\
  Self-Attention Heads & 12  \\
  Activation Function & SwiGLU \citep{shazeer2020glu} \\
  Normalization Layer & Adaptive RMSNorm \citep{zhang2019root, peebles2022scalable}  \\
  Optimizer & AdamW \citep{loshchilov2018decoupled}\\
  Learning Rate & {1e-3} \\
  ($\beta_1$, $\beta_2$) & {(0.9, 0.999)} \\
  Batch Size &  256 \\
  Warmup Steps & 1000 \\
  Learning Rate Schedule & Cosine Decay \\
  Weight Decay & {.1} \\
  Gradient Clipping & {1.0}\\
  Batch Size & 256 \\
Training Noise Schedule & Adaptive \citep{kingma2023understanding} \\
Sampling Noise Schedule & Cosine \citep{improved-ddpm-nichol21a, hoogeboom2023simple} \\
  Training Steps & {250k} \\
\bottomrule
\end{tabular}
\end{table*}

\section{Additional Composition Results}
As specified in the main paper, \method{} is naturally suited for simultaneously controlling multiple attributes. \autoref{app:composition} presents additional qualitative results for compositional control. From the table, we see that the instances satisfies both control attributes. 

\begin{table*}[htbp]
\small
\centering
\resizebox{\linewidth}{!}{
\begin{tabular}{cccc}
\toprule
 Guidance & Prefix & Continuation \\
  
\midrule
  Low Toxicity & \parbox{.35\linewidth}{ She made out with him, took sexy pictures in a photo booth, and watched him} & \parbox{.35\linewidth}{ have fun with his friends. She shared the photos on her private Facebook page, where a family...} \\
  \midrule
  Positive Sentiment & Cycle, published by the CTC, is running & \parbox{.35\linewidth}{  its 10th edition and it is getting better every time I see the contents! It's also very...} \\
  \midrule
  Negative Sentiment & Cycle, published by the CTC, is running & \parbox{.35\linewidth}{  its ‘news’ section, with no substance at all and zero interest in the subject it...} \\
\bottomrule
\end{tabular}
}
\caption{Qualitative examples of guided generation.}\label{tab:qualitative}

\end{table*}

\label{app:composition}
\begin{table*}[htbp]
\small
\centering
\resizebox{\linewidth}{!}{
\begin{tabular}{cccc}
\toprule
 Topic & Sentiment & Prefix & Continuation \\
  
\midrule
  Sports & Negative & The person that makes such decisions & \parbox{.35\linewidth}{and coaches the team, is really just a bad sports person. That’s all there ever was to it. I think we are going down...} \\
  \midrule
  Business & Negative & I realized this 10 years & \parbox{.35\linewidth}{ ago when I started running my own business and it was very clear to me that Sunland was really not making any money...} \\
  \midrule
  Sci/ Tech & Positive & And that’s pretty much the & \parbox{.35\linewidth}{ best part about this site – you can get all sorts of great technical info with just a couple clicks...} \\
\bottomrule
\end{tabular}
}
\caption{Additional Qualitative examples of compositional control.\label{tab:compositional}}

\end{table*}

\end{document}